\documentclass[letterpaper, 10 pt, conference]{ieeeconf}  

\def\ie{{\textit{ i.e.}}}
\def\eg{{\textit{ e.g.}}}
\def\etal{{\textit{et al.}}}

\newcommand{\figref}[1]{Fig. \ref{#1}}
\newcommand{\tabref}[1]{Tab. \ref{#1}}
\newcommand{\equref}[1]{(\ref{#1})}
\newcommand{\secref}[1]{Section \ref{#1}}

\newcommand{\mc}[1]{\mathcal{#1}}

\newcommand{\bs}[1]{\boldsymbol{\texttt{#1}}}

\usepackage{epstopdf}
\usepackage{amsmath}
\usepackage{graphicx}
\usepackage{cite}
\usepackage{url}
\usepackage{ragged2e}
\usepackage{booktabs}
\usepackage{color}
\usepackage{multirow}
\usepackage{bm}
\usepackage{bbm}
\usepackage[misc]{ifsym}

\usepackage{multirow}
\usepackage{float}

\usepackage{amssymb}

\IEEEoverridecommandlockouts                              

\title{\LARGE \bf
Focus on BEV: Self-calibrated Cycle View Transformation \\ for Monocular Birds-Eye-View Segmentation
}

\author{Jiawei Zhao$^{1,\dagger}$, Qixing Jiang$^{1,\dagger}$, Xuede Li$^{1}$, Junfeng Luo$^{1}$
\thanks{$^{\dagger}$ equal contribution.}
\thanks{$^{1}$Vision Intelligence Department (VID), Meituan, Beijing, China, \{zhaojiawei12, jiangqixing, lixuede, luojunfeng\}@meituan.com
        }%
}

\begin{document}

\maketitle
\thispagestyle{empty}
\pagestyle{empty}

\begin{abstract}
Birds-Eye-View (BEV) segmentation aims to establish a spatial mapping from the perspective view to the top view and estimate the semantic maps from monocular images. Recent studies have encountered difficulties in view transformation due to the disruption of BEV-agnostic features in image space. To tackle this issue, we propose a novel FocusBEV framework consisting of $(i)$ a self-calibrated cross view transformation module to suppress the BEV-agnostic image areas and focus on the BEV-relevant areas in the view transformation stage, $(ii)$ a plug-and-play ego-motion-based temporal fusion module to exploit the spatiotemporal structure consistency in BEV space with a memory bank, and $(iii)$ an occupancy-agnostic IoU loss to mitigate both semantic and positional uncertainties. Experimental evidence demonstrates that our approach achieves new state-of-the-art on two popular benchmarks,\ie, 29.2\% mIoU on nuScenes and 35.2\% mIoU on Argoverse. 
\end{abstract}

\section{INTRODUCTION}

Monocular birds-eye-view (BEV) segmentation aims to parse the road layout and locate the occupancy of traffic agents on the BEV plane from front-view monocular images, which serves as an important prerequisite for many automation applications, such as autonomous driving \cite{2022visionbev,2022delving}, robot navigation \cite{2020VPN,ross2022bev}, map reconstruction \cite{2022hdmapnet,2023vectormapnet,liao2022maptr,peng2023bevsegformer}.
Different from the well-learned 2D segmentation tasks \cite{2017deeplab,2020HRNet,2021segformer}, the key issue in BEV segmentation tasks lies in transforming the perspective view (PV) space to the BEV space and inferring the semantic occupancy map on the BEV plane, which is inherently ill-posed.

One intuitive solution for transforming different views is to apply geometric mapping mechanisms \cite{ipm,2012IPMSeg,2020LSS}. However, these pioneer attempts are plagued by mapping distortion due to the inherent limitations in the ill-posed view transformation. To mitigate this problem, recent studies \cite{2020VPN,2020PON,2022TIM} employ MLP or Transformer to implicitly build mapping correspondence between image features and BEV features with geometry priors. However, these methods typically transform the BEV plane from the entire image regions, which inherently include numerous BEV-agnostic areas that disrupt the view transformation. Besides, the spatiotemporal consistency across multiple frames remains less explored.

\begin{figure}
	\begin{center}
		\includegraphics[width=.95\linewidth]{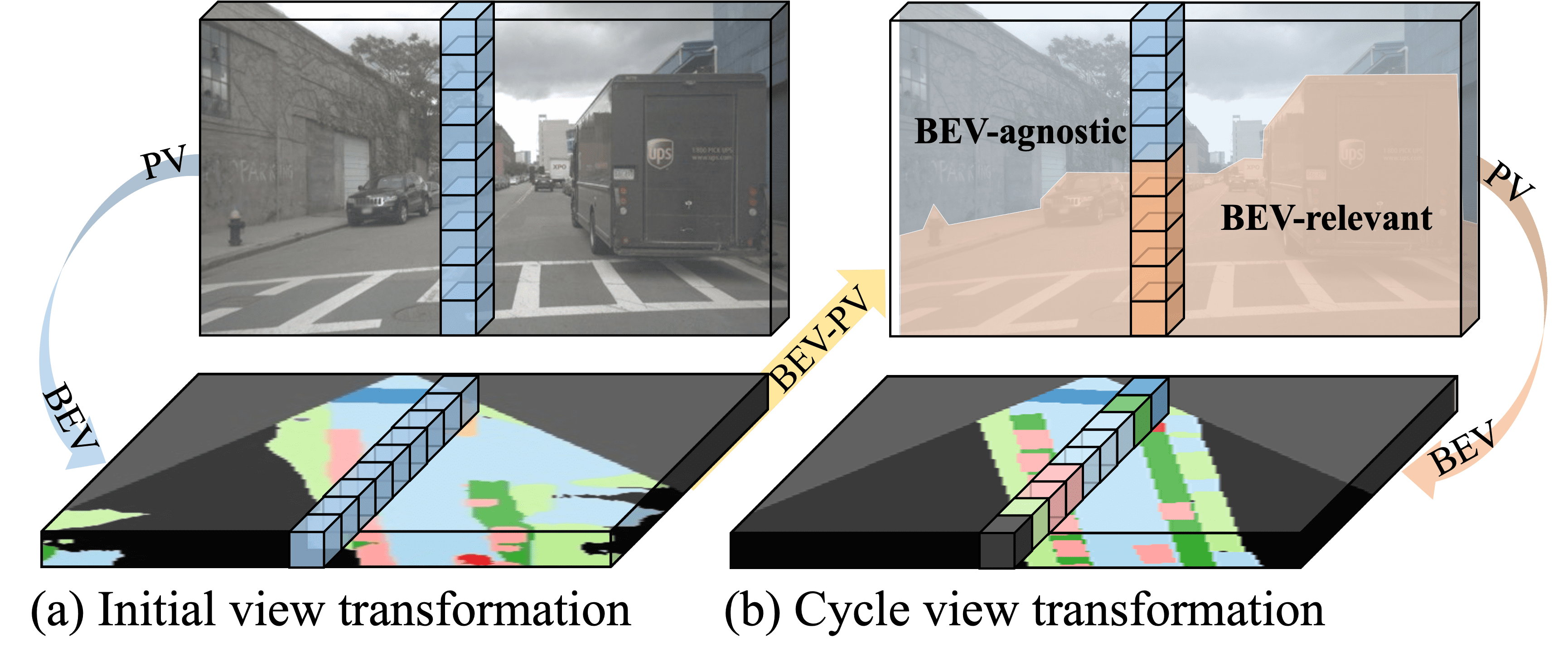}
		\caption{Motivation of the proposed self-calibrated cycle view transformation. (a) Initial PV-BEV transformation is disrupted by BEV-agnostic areas (\eg, sky and buildings), whereas (b) the cycle view transformation could suppress these BEV-agnostic areas and concentrate on BEV-relevant areas in a self-calibrated manner.
		}\label{fig:motivation}
	\end{center}
\end{figure}

To address the aforementioned concerns, we propose a FocusBEV framework to suppress the BEV-agnostic image regions and focus on the BEV-relevant image regions with cycle view transformation in a self-calibrated manner. 
As shown in \figref{fig:motivation} (a), in the initial PV-BEV transformation stage, the obtained BEV features contain lots of noise outside the road regions due to the disruption caused by numerous BEV-agnostic areas in image features. 
To suppress the traffic-agnostic areas as well as emphasize the BEV-relevant areas in image features, we propose a novel self-calibrated scheme via cycle view transformation (\ie, BEV-PV-BEV) in \figref{fig:motivation} (b), which first adopts a BEV-PV transformer on the initial obtained BEV features to implicitly generate image features that focus on BEV-relevant areas with the BEV structure guidance, and then employs a PV-BEV transformer on the generated image features to transform the BEV-relevant areas into BEV plane to obtain the calibrated BEV features with less disruption.
To process the image sequence, we propose a plug-and-play ego-motion-based temporal fusion module to focus on the spatiotemporal structure in BEV space, which aligns and aggregates multiple history features from the BEV memory bank. 
Besides, we introduce an occupancy-agnostic IoU loss to mitigate both semantic and positional uncertainties.
In the end, our proposed approach achieves the state-of-the-art results on two most popular benchmarks, \ie, nuScenes\cite{2020nuscenesdataset} and Argoverse\cite{2019argoversedataset}.

\begin{figure*}[htb]
	\begin{center}
		\includegraphics[width=1\textwidth]{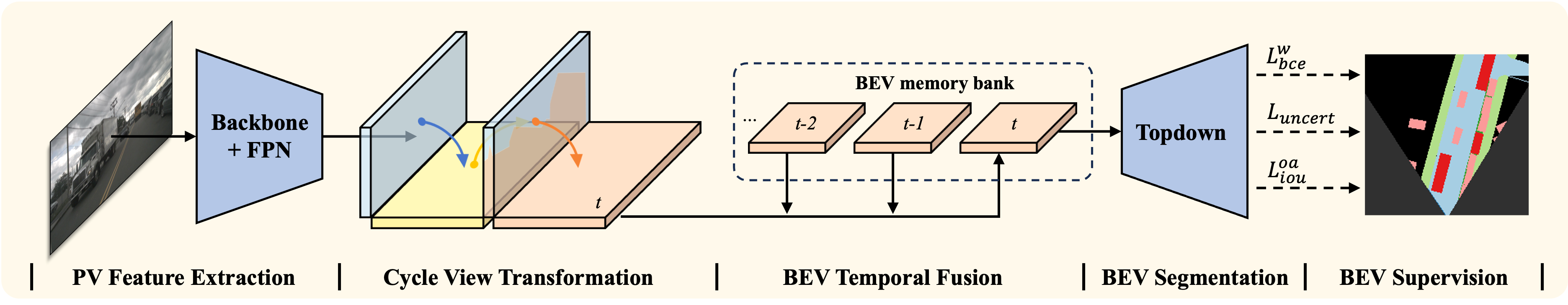}
		\caption{The overall pipeline of our proposed FocusBEV framework. The backbone and FPN extract PV features. The cycle view transformation then transforms the spatial features of PV space to the BEV space using cyclical mapping. The BEV temporal fusion aligns the history BEV features and aggregates them spatiotemporally with the current frame. The top-down network upsamples BEV features to predict a semantic occupancy map. 
		}\label{fig:pipeline}
	\end{center}
\end{figure*}

In summary, our contribution is three-fold: 1) We propose a novel FocusBEV framework with a self-calibrated cycle view transformation module, which suppresses the BEV-agnostic regions and focuses on the BEV-relevant regions via a cycle transformation scheme. 2) A plug-and-play ego-motioned-based temporal fusion module is constructed to align and aggregate the multiple history BEV features within the spatiotemporal space. 3) An occupancy-agnostic IoU loss is introduced to address both semantic and positional uncertainties on the BEV plane.

\section{RELATED WORK}
The transformation from perspective view (PV) to birds-eye-view (BEV) is a crucial
problem in BEV segmentation tasks. Based on the varying strategies of view transformation, most monocular BEV segmentation methods can be broadly categorized into three groups: geometry-based methods, MLP-based methods, and Transformer-based methods.

\textbf{Geometry-based BEV Segmentation.} 
 Previous pioneer works \cite{2012IPMSeg,2020bevstitch} tend to typically employ mathematical mapping to convert front-view images to top-view images using Inverse Perspective Mapping (IPM) \cite{ipm}.
For example, Sengupta \etal \cite{2012IPMSeg} project the semantic segmentation predictions of the image plane to the BEV plane via a homography map. 
Can \etal \cite{2020bevstitch} propose a two-branch network to segment the static layout and dynamic object footprints in the image plane, then warp them to the BEV plane by IPM.
However, due to IPM's strong flat-ground assumption, these methods often fail to locate and distinguish objects above the ground, such as pedestrians and bicycles. 
To avoid the mapping distortion caused by IPM, recent studies leverage additional depth to lift the 2D space to 3D. 
For example, Philion \etal \cite{2020LSS} lift 2D pixels to 3D points with the learned depth distribution and splat to the BEV plane. Similarly, Dwivedi \etal \cite{dwivedi2021bird} apply estimated depth to transform 2D features to BEV features. 

\textbf{MLP-based BEV Segmentation.} Some works \cite{2019VED,mani2020monolayout,2020PON,2021STAS,2021PYVA,2023HFT}
tend to apply Multi-Layer Perceptron (MLP) to implicitly exploit the view transformation between image space and BEV space. Pan \etal \cite{2020VPN} and Li \etal \cite{2022hdmapnet} adopt a two-layer MLP on the flattened image features, thereby transforming PV features to BEV features. Roddick \etal \cite{2020PON} and Saha \etal \cite{2021STAS} propose column-wise MLP to collapse the height axis of pyramid image features and expand along the depth axis of BEV features.
Yang \etal \cite{2021PYVA} propose a cross-view transformer to leverage the cycle consistency between views.
Furthermore, Zhou \etal \cite{2023HFT} design a two-branch network to learn a hybrid feature transformation with the geometric information and global context respectively.

\textbf{Transformer-based BEV Segmentation.}
Some works \cite{2022TIM,2022gitnet,zhou2022CVT,pan2023baeformer} devote to mapping the image space to BEV space with transformer structure \cite{vaswani2017attention,2020DETR}. For example, Saha \etal \cite{2022TIM} first introduce an encoder-decoder transformer to transform the vertical scanlines in the image plane to the polar rays on the BEV plane. Gong \etal \cite{2022gitnet} propose a two-stage view transformation framework, employing an encoder-decoder transformer to enhance the initial geometry-aware BEV features with column-wise attention.
However, these methods often inadvertently overlook the potential disruption of BEV-agnostic features in view transformation.

\section{METHOD}
\subsection{Overview}
In this section, we introduce a novel FocusBEV framework that incorporates a self-calibrated cycle view transformation module and an ego-motion-based temporal fusion module for front-view monocular BEV segmentation, focusing on BEV-relevant spatial mapping and spatiotemporal aggregation in BEV space respectively in \figref{fig:pipeline}. 
Given the reference image $I^{t} \in \mathbb{R}^{H \times W \times 3}$ as input, the objective of BEV segmentation tasks is to predict the BEV semantic occupancy map $M^{t}\in \mathbb{R}^{Z \times X \times N_{c}}$ in the ego camera coordinate, where $H$ and $W$ denote the height and width of images, $Z$ and $X$ denote the spatial dimensions of BEV maps, $N_{c}$ is the number of semantic categories.

\begin{figure*}
	\begin{center}
		\includegraphics[width=\linewidth]{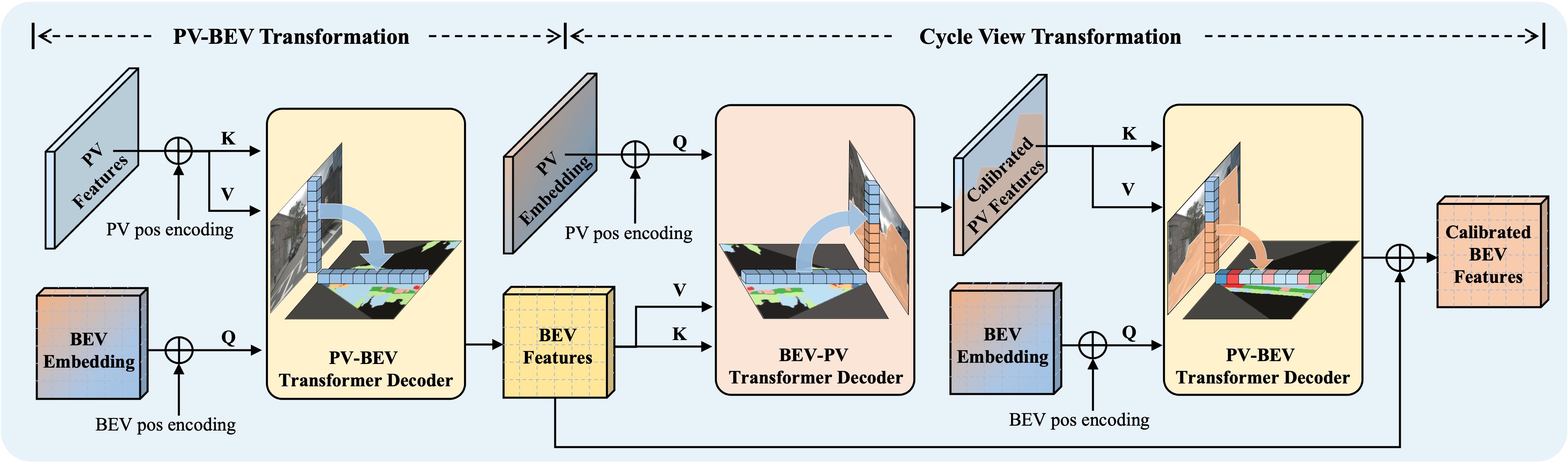}
		\caption{Our proposed self-calibrated cycle view transformation module consists of two stages: the PV-BEV transformation and the cycle view transformation. 
		}\label{fig:pv2bev}
	\end{center}
\end{figure*}

Specifically, we apply a shared backbone and FPN \cite{2017FPN} on the reference frame $I^{t}$ to extract the multi-level PV features $\{ F^{t}_{1}, ..., F^{t}_{5} \}$. To concentrate on BEV-relevant areas during view transformation, we perform a self-calibrated cycle transformation from the PV space to BEV space to obtain the calibrated BEV features $B^{t}_{calib}$ as depicted in \secref{BCVT}. To aggregate the spatiotemporal contextual information across multiple frames, the history BEV features $\{B^{t-N_{his}}_{calib}, ..., B^{t-1}_{calib} \}$ are retrieved to be aligned with ego-motion and aggregated to obtain the enhanced BEV feature $B^{t}_{temp}$, where $N_{his}$ is the length of temporal aggregation window. To predict the semantic maps, we employ a top-down network composed of two residual bottleneck layers for upsampling and segmentation, following \cite{2020PON}. Finally, we apply the weighted binary cross-entropy loss and uncertain loss following \cite{2020PON} and introduce an auxiliary occupancy-agnostic IoU loss to supervise the final BEV predictions. 


\subsection{Self-calibrated Cycle View Transformation} 
\label{BCVT}
The crucial problem in BEV segmentation is to select the BEV-relevant feature regions in the image space and build an accurate semantic mapping between the image space and BEV space. 
To address this, we incorporate the column-wise Transformer decoder into view transformation to build the global spatial mapping between different views and propose a novel cycle view transformation scheme to suppress the BEV-agnostic features, thereby enhancing the BEV representation. 

\textbf{Revisiting Transformer Decoder for View Transformation.}
In the field of natural language processing, the conventional decoder in transformers \cite{vaswani2017attention} build global translation relationships between 1D language sentences of varying lengths. Inspired by language processing, 2D images could be collapsed into 1D sequence with positional encoding by flattening pixels, and geometric view transformation could be treated as the sequence-to-sequence translating problem \cite{2022TIM,2020DETR}. Considering the rough geometric correspondence between Cartesian columns in the image plane and polar rays in the BEV plane \cite{2020PON, 2022TIM}, we apply the transformer decoder on column-wise pixels rather than 
all image pixels for view transformation, which could effectively mitigate the uncertainty of view transformation and reduce the complexity of attention.


Specifically, to transform the given $i$-$th$ level image features $F_{i} \in \mathbb{R}^{C_{T} \times H_{i} \times W_{i} }$ into the BEV features $B_{i}\in \mathbb{R}^{C_{T} \times Z_{i} \times X}$ in $i$-$th$ depth range, we first construct the $query$ with a grid-shape learnable embedding $E_{i} \in \mathbb{R}^{C_{T} \times Z_{i} \times X}$, construct the $key$ and $value$ with $F_{i}$ for column-wise cross attention modules:
 \begin{equation}
 \begin{split}
 \mc{Q}(E_{i}) &= (\mc{R}^{c}(E_{i}) + \mc{E}(\mc{R}^{c}(E_{i})))W_{Q}, \\
\mc{K}(F_{i}) &= (\mc{R}^{c}(F_{i}) + \mc{E}(\mc{R}^{c}(F_{i})))W_{K}, \\
\mc{V}(F_{i}) &= (\mc{R}^{c}(F_{i}) +)W_{V}, 
 \end{split}
 \end{equation}
 where $\mc{R}^{c}(\cdot)$ denotes the column-wise reshape operation to collapse along the height axis, $\mc{E}(\cdot)$ denotes the learnable position encoding operation, $W_{\{Q, K, V\}}$ denote the learnable weights of $query$, $key$ and $value$ projection. 
To effectively transform the multi-scale image features $\{F^{t}_{1}, ..., \mc{F}^{t}_{5} \}$ conducted by FPN to BEV features $B^{t}$, we apply a column-wise decoder-only transformer structure as PV-BEV view transformer $\mc{D}^{PV-BEV}$, which consists of $N_{dec}$ layers of Multi-Head Cross-Attention modules (MHCA) and Multi-Layer Perception blocks (MLP): 
 \begin{equation}
 \label{eq:pv2polar}
 \begin{split}
E_{i}^{'} &= \delta(\bs{MHCA}(\mc{Q}(E_{i}), \mc{K}(F_{i}), \mc{V}(F_{i})) + \mc{Q}(E_{i})), \\
P_{i} &= \mc{D}^{PV-BEV}_{i}(E_{i}, F_{i}) = \mc{R}^{e}(\delta(\bs{MLP}(E_{i}^{'})+ E_{i}^{'})), \\
 \end{split}
 \end{equation}
where $\delta(\cdot)$ denotes the Layer Normalization \cite{ba2016LN}, $\mc{R}^{e}(\cdot)$ denotes the ray-wise reshape operation for expansion along the depth axis, $P_{i} \in \mathbb{R}^{C_{T} \times Z_{i} \times W_{i}}$ denotes the polar BEV features in the polar coordinate system. Following \cite{2020PON}, we resample the polar BEV features to Cartesian coordinates using the known camera focal length $f$ and horizontal offset $u_{0}$ and concatenate along the depth axis to obtain the final BEV features $B$:
 \begin{equation}
 \label{eq:polar2cart}
 \begin{split}
B_{i} &= \mc{S}(P_{i}, \frac{f}{d_i}, \frac{u_{0}}{d_i}) \in \mathbb{R}^{C_{T} \times Z_{i} \times X}, \\
B &= \bs{cat}(\{B_{1}, B_{2}, B_{3}, B_{4}, B_{5}\}) \in \mathbb{R}^{C_{T} \times Z \times X},
 \end{split}
 \end{equation}
where $d_{i}=2^{i+2}, i\in\{1, ..., 5\}$ is the downsampling factors.

\textbf{Focusing on BEV with Cycle View Transformer.}
A large portion of regions (\eg, sky and buildings) in the image are irrelevant to the road layout and traffic participants, which motivates us to suppress these BEV-agnostic features and focus on the BEV-relevant features in view transformation.
Toward this, we propose a self-calibrated scheme to suppress the BEV-agnostic features and focus the BEV-relevant features with cycle view transformation as illustrated in \figref{fig:pv2bev}. We first apply a column-wise PV-BEV transformation aforementioned to obtain the initial polar BEV features,  then a cycle BEV-PV-BEV transformation is adopted to obtain the BEV-focusing PV features and calibrate the view transformation with the implicit BEV-focusing guidance.

In the first PV-BEV transformation stage, we apply a column-wise PV-BEV view transformer on the referenced image features $F_{i}$ and a grid-shape learnable embedding $E_{i}^{BEV}$ to obtain the initial polar BEV features $P_{i}$ as illustrated in Eq. \equref{eq:pv2polar}:
 \begin{equation}
 \begin{split}
P_{i} &= \mc{D}^{PV-BEV}_{i}(E_{i}^{BEV}, F_{i}). 
 \end{split}
 \end{equation}
The initial polar BEV features contain rough BEV content with lots of noise due to the BEV-agnostic disruption during the view transformation. 

In the second BEV-PV-BEV cycle transformation stage, we first employ a column-wise BEV-PV view transformer on the initial polar BEV features $P_{i}$ and an image-shaped learning embedding $E_{i}^{PV}$ to obtain the BEV-focusing PV features $F_{i}^{calib}$ using an implicit self-calibrated scheme:
  \begin{equation}
 \begin{split}
F^{calib}_{i} &= \mc{D}^{BEV-PV}_{i}(E^{PV}_{i}, P_{i}), 
 \end{split}
 \end{equation}
which could effectively suppress the BEV-agnostic contextual information and enhance the BEV-relevant structure in the original PV features via an inverse view transformation. With this BEV-focusing guidance, we further utilize the weight-shared PV-BEV view transformer with a grid-shaped learnable embedding $\Tilde{E}_{i}^{BEV}$ as well as a residual connection with the initial polar BEV features, to obtain the calibrated polar BEV features:
 \begin{equation}
 \begin{split}
P^{calib}_{i} &= \mc{D}^{PV-BEV}_{i}(\Tilde{E}^{BEV}_{i}, F^{calib}_{i}) + P_{i}.
 \end{split}
 \end{equation}
Finally, we concatenate the multi-range polar BEV features $\{P_{1}^{calib}, ..., P_{5}^{calib} \}$ transformed from multi-scale image features $\{F_{1}, ..., F_{5} \}$ and resample into the Cartesian coordinates to generate the calibrated BEV features $B_{calib}$, following the Eq. \equref{eq:polar2cart}.

\subsection{Ego-motion-based Temporal Fusion}
\label{TempFusion}

\begin{figure}[t]
	\begin{center}
		\includegraphics[width=\linewidth]{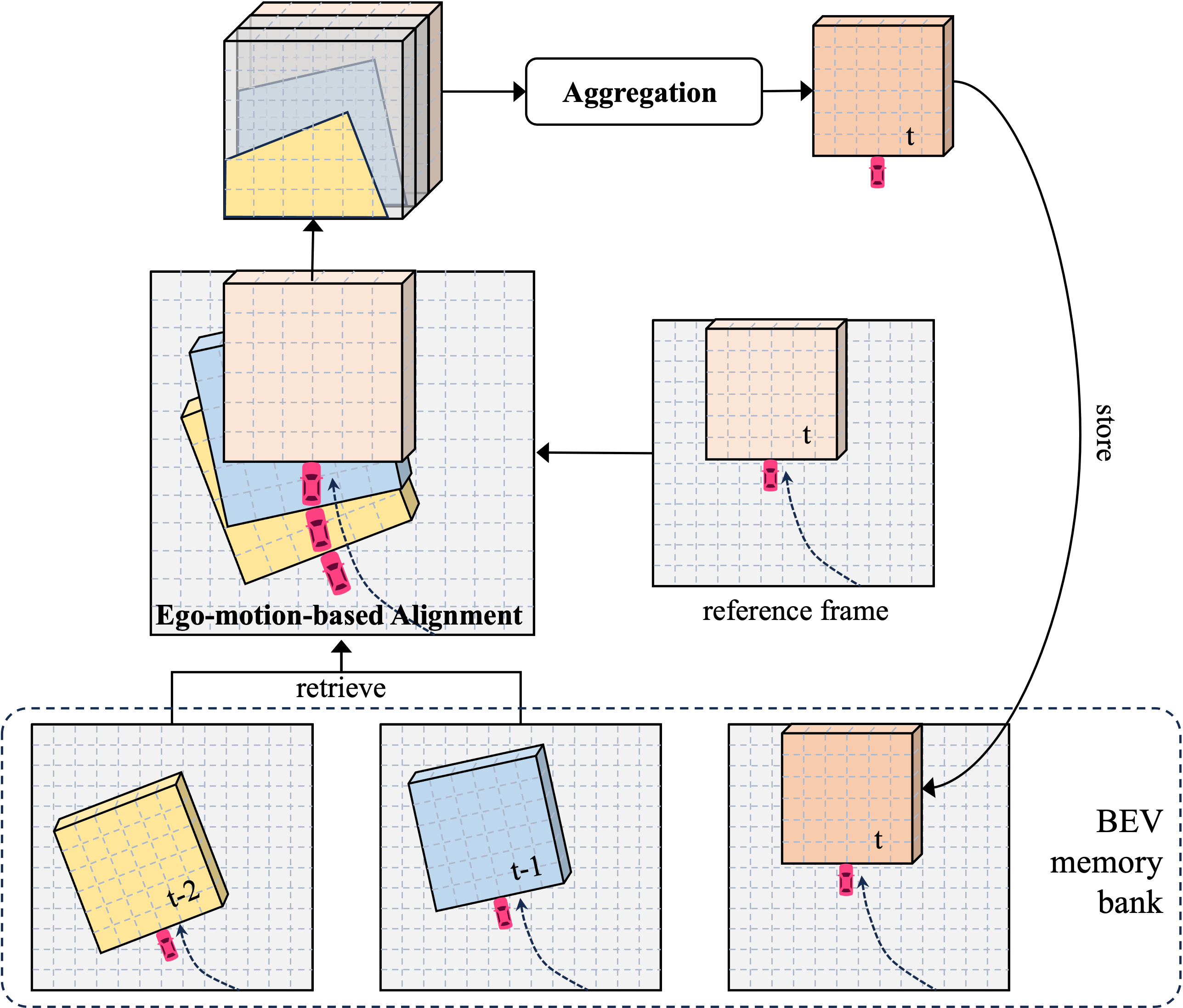}
		\caption{Our proposed ego-motion-based temporal fusion module. The history BEV features are aligned with the reference frame using ego-motion. These features are then stacked and aggregated to enhance the spatiotemporal contextual information of the reference frame.
		}\label{fig:temporal_module}
	\end{center}
\end{figure}

To mitigate the intermittent occlusions inherent in monocular BEV segmentation, we adopt ego-motion-based alignment and temporal aggregation as depicted in \figref{fig:temporal_module} to exploit the spatiotemporal structure consistency in BEV space with a memory bank, which contains BEV history features and corresponding ego-motion information. 
With these explicit alignment and aggregation performed, our proposed temporal fusion module could accelerate the convergence and effectively mitigate the disruption of occlusions among multiple frames and benefit from the spatiotemporal consistent structure of static categories (\eg, crosswalk and drivable areas) as well as the motion contextual information of dynamic categories (\eg, cars and pedestrians). 

\textbf{Ego-motion-based Alignment.}
Given the reference BEV feature $B_{calib}^t$ obtained by our proposed view transformation module, we retrieve multiple history features $\{B_{calib}^{t-N_{his}}, ..., B_{calib}^{t-1} \}$ from the memory bank and employ ego-motion information to align $i$-$th$ history BEV feature to the reference pose through typical affine transformations:
\begin{equation}
\label{eq:align_prev_bev}
B_{align}^{t-i} = 
 \left\{\begin{array}{ll}
        \- \mc{m}(\mc{r}(B_{calib}^{t-i}, r^t_{t-i}), m^t_{t-i}) &\mathrm{if\ s_t = s_{t-i}} \\
        \-B_{calib}^{t} &\mathrm{otherwise}
       \end{array}
 \right. ,
\end{equation}
where $\mc{r}(\cdot)$ represents the rotation transformation with camera rotation matrix $r_{t-i}^{t}$ from frame $t-i$ to frame $t$, $\mc{m}(\cdot)$ represents the translation transformation by ego motion $m_{t-i}^{t}$ from frame $t-i$ to frame $t$. Specially, we replace the history feature with the reference feature if the scene of timestamp $s_{t-i}$ is inconsistent with $s_t$. 
In the end, we obtain the spatiotemporal aligned features $\{B_{align}^{t-N_{his}}, ..., B_{align}^{t-1}, B_{calib}^{t} \}$ to explicitly compensate the occlusion in the BEV space.


\textbf{Temporal Aggregation.}
To aggregate the aligned history features aforementioned with the reference BEV feature $B_{calib}^{t}$, we concatenate them along the channel dimension and aggregate their spatiotemporal structure consistency to obtain the final spatiotemporal enhanced feature $B^{t}_{temp}$ in a learnable manner:
\begin{equation}
\label{eq:origin_fl}
B^{t}_{temp} = \phi(\bs{cat}(B_{align}^{t-N_{his}}, ..., B_{align}^{t-1}, B_{calib}^{t} )),
\end{equation}
where $\phi$ denotes a $1 \times 1$ convolution layer to enhance the spatial structure and reduce channel dimensions. Notably, the history BEV features only participate in the forward-propagation but not the back-propagation during both training and testing.






\begin{table*}[t]
\centering{
\setlength\tabcolsep{4pt}

\caption{Comparisons with state-of-the-art methods on the NuScenes dataset. }
\label{nuscene benchmark}
\renewcommand{\arraystretch}{1.1}
\begin{tabular}{c|cccc|cccccccccc||c}
\hline
\multicolumn{1}{c|}{\multirow{2}{*}{Methods}} & \multicolumn{4}{c|}{Layout}            & \multicolumn{10}{c||}{Object}                                                        & \multirow{2}{*}{mIoU} \\
\multicolumn{1}{c|}{}                         & Drivable & Crossing & Walkway & Carpark & Car  & Truck & Bus  & Trailer & C.V. & Ped. & Motor & Bike & Cone & Barrier &                       \\ 
\hline  \hline
VED \cite{2019VED}                            & 54.7     & 12.0    & 20.7    & 13.5    & 8.8  & 0.2   & 0.0  & 7.4     & 0.0       & 0.0  & 0.0   & 0.0     & 0.0  & 4.0     & 8.7                   \\
VPN  \cite{2020VPN}                           & 58.0     & 27.3    & 29.4    & 12.9    & 25.5 & 17.3  & 20.0 & 16.6    & 4.9       & 7.1  & 5.6   & 4.4     & 4.6  & 10.8    & 17.5                  \\
PYVA \cite{2021PYVA}                          & 56.2     & 26.4    & 32.2    & 21.3    & 19.3 & 13.2  & 21.4 & 12.5    & 7.4       & 4.2  & 3.5   & 4.3     & 2.0  & 6.3     & 16.4                  \\
LSS  \cite{2020LSS}                           & 55.9     & 31.3    & 34.4    & 23.7    & 27.3 & 16.8  & 27.3 & 17.0    & 9.2       & 6.8  & 6.6   & 6.3     & 4.2  & 9.6     & 19.7                  \\
PON  \cite{2020PON}                          & 60.4     & 28.0    & 31.0    & 18.4    & 24.7 & 16.8  & 20.8 & 16.6    & 12.3      & 8.2  & 7.0   & 9.4     & 5.7  & 8.1     & 19.1                  \\
HFT  \cite{2023HFT}                          & 56.3     & 36.2    & 35.8    & 23.8    & 31.2 & 19.7  & 28.5 & 19.1    & 6.7       & 8.5  & 12.0  & 12.4    & 6.7  & 11.4    & 22.1                  \\
BEVStitch  \cite{2020bevstitch}    & 74.9 & 31.6 & 38.1 & 36.8 & 36.5 & 18.1 & 27.4 & 17.6 & 4.4 & 6.4 & 8.6 & 7.9 & 4.9 & 17.2    & 23.6                  \\
STA-S  \cite{2021STAS}                       & 70.7     & 31.1    & 32.4    & 33.5    & 36.0 & 22.8  & 29.2 & 13.6    & 12.1      & 8.6  & 8.0   & 12.1    & 6.9  & 14.2    & 23.7                  \\
TIM  \cite{2022TIM}                         & 74.5     & 36.6    & 35.9    & 31.3    & 39.7 & \textbf{26.3}  & 32.8 & 13.9    & \textbf{14.2}      & 9.5  & 7.6   & 14.7    & 7.6  & 14.7    & 25.7                  \\
DiffBEV  \cite{2023DiffBEV}                & 65.4     & 41.3    & 41.1    & 28.4    & 38.9 & 23.1  & 33.7 & 21.1    & 8.4       & 9.6  & \textbf{14.4}  & 13.2    & 7.5  & 16.7    & 25.9                  \\ 
\hline 
Ours                                     &   \textbf{75.3}     &    \textbf{42.6}     &  \textbf{42.0}  & \textbf{42.1}     &   \textbf{43.2}      &   26.2   &   \textbf{37.8}    &  \textbf{27.2}    &     2.5    &     \textbf{10.3}      & 13.6   &   \textbf{13.4}    &    \textbf{7.8}     &   \textbf{24.1}        & \textbf{29.2}               \\
\hline
\end{tabular}
}
\end{table*}


\subsection{Learning Objective}
\label{loss}
With the semantic map s'dprediction obtained from a top-down network, we first adopt a weighted binary cross-entropy loss $\mc{L}_{bce}^{w}$ on visible regions and an uncertain loss $\mc{L}_{uncert}$ on occluded regions, following \cite{2020PON}. However, the BCE loss often results in noisy predictions that fail to form a holistic object. To address both semantic and positional uncertainties, we propose an occupancy-agnostic IoU (Intersection over Union) loss $\mc{L}_{iou}^{oa}$ for all regions:
\begin{equation} \label{eq:iou loss}
\begin{split}
\mc{L}_{iou}^{oa} = 1-\sum_{k=1}^{N_c}\frac{1}{N_c}\frac{\sum_{i=1}^{Z}\sum_{j=1}^{X} (y^k_{ij}\times p^k_{ij})+1}{\sum_{i=1}^{Z}\sum_{j=1}^{X}(y^k_{ij}+p^k_{ij}-y^k_{ij}\times p^k_{ij})+1},
\end{split}
\end{equation}
where $p^k, y^k \in \mathbb{R}^{Z\times X\times 1}$ are the predicted map and corresponding groudtruth of $k$-$th$ category on all regions. The final learning objective $L_{sum}$ can be formulated as:
\begin{equation} \label{eq:sum loss}
\begin{split}
\mc{L}_{sum}  = \mc{L}^{w}_{bce} + \alpha \mc{L}_{uncert} + \beta \mc{L}_{iou}^{oa},
\end{split}
\end{equation}
where $\alpha$ denotes the weight of uncertain loss and $\beta$ denotes the weight of occupancy-agnostic IoU loss. Here we set $\alpha$ as 0.001 and $\beta$ as 0.01.

\begin{table}[]
\centering{
\setlength\tabcolsep{4pt}

\caption{Ablation study on the nuScenes dataset.}
\label{ablation}
\renewcommand{\arraystretch}{1.2}
\begin{tabular}{ccc|ccc}
\hline
\multirow{2}{*}{SCVT} & \multirow{2}{*}{$\mc{L}_{iou}^{oa}$} & \multirow{2}{*}{ETF}  & \multicolumn{3}{c}{mIoU} \\
               &                     &                                    & Layout  & Object & Total \\ \hline
                &                    &                                    &  38.2       &   12.1     & 19.5  \\
                &                    & \checkmark                        &   40.4      &   13.1   & 20.9  \\
               & \checkmark           &                                &     40.4    &   12.9     & 20.3     \\ \hline
\checkmark      &                   &                                    &   50.2      &    19.6    & 28.4     \\ 
\checkmark       &      \checkmark           &                         &    49.6    &    20.5    & 28.8     \\
\checkmark        & \checkmark          & \checkmark                   & \textbf{50.5}    &   \textbf{20.6}   & \textbf{29.2}  \\ \hline
\end{tabular}
}
\end{table}

\section{EXPERIMENTS}
\subsection{Datasets and Evaluation Metrics}


\textbf{NuScenes Benchmark.} NuScenes \cite{2020nuscenesdataset} is a large-scale autonomous driving dataset consisting of 1,000 sequences from 6 surrounding cameras. Following \cite{2020PON, 2022TIM, 2023HFT}, we only use the images captured by the front camera and split the dataset into a training set with 28,008 images and a validation set of 5,981 images, which contains 4 static categories and 10 dynamic categories for the BEV segmentation task. 

\textbf{Argoverse Benchmark.} Argoverse \cite{2019argoversedataset} is another popular benchmark for BEV segmentation. It contains 6723 images in 65 training sequences and 2418 images in 24 validation sequences with 7 dynamic categories and 1 static category. 

For both datasets, the ground truth of BEV semantic maps expands from 1m to 50m in front of the camera (i.e., along the z-direction) and 25m to either side (i.e., along the x-direction).

\textbf{Evaluation Metrics.} To quantitatively evaluate the performance of our approach and state-of-the-art methods, we employ the IoU (Intersection over Union) accuracy and mIoU as evaluation metrics. Given the predicted semantic map, we conduct a binary map for each class based on a threshold of 0.5, following \cite{2020PON}. Both the predicted maps and ground truth are resized to $196 \times 200$ pixels to be comparable with \cite{2020PON,2023HFT,2022TIM}.

\begin{table}[]
\centering{
\setlength\tabcolsep{4pt}

\caption{Ablation study for transformer decoder layers in self-calibrated cross view transformation. }
\label{ablation_scvt}
\renewcommand{\arraystretch}{1.2}
\begin{tabular}{c|ccccccc}
\hline
$N_{dec}$ & 0 &  1 & 2 & 3 & 4 & 5 & 6   \\ \hline
mIoU   & 25.7 & 28.5 & \textbf{29.2}  & 28.6  & 28.7 & 27.9 & 26.9 \\ \hline
\end{tabular}
}
\end{table}

\begin{table}[]
\centering{
\setlength\tabcolsep{4pt}

\caption{Ablation study for ego-motion-based temporal fusion. }
\label{ablation_etf}
\renewcommand{\arraystretch}{1.2}
\begin{tabular}{c|ccccccc}
\hline
$N_{his}$ & 0 &  1 & 2 & 3 & 4 & 5  \\ \hline
Layout  & 49.6 & 50.0  & 50.5  & 50.5  & 50.3  &  \textbf{50.9}  \\ 
Object  & 20.5 & 19.9  & \textbf{20.6}  & 20.3  & 19.7  &  20.4   \\ 
Total   & 28.8 & 28.5  & \textbf{29.2}  & 29.0  & 28.5  & 29.1  \\ \hline
\end{tabular}
}
\end{table}

\subsection{Implementation Details}
We adopt ResNe-50 \cite{2016ResNet} pre-trained on ImageNet \cite{deng2009imagenet} as our backbone.
In the training stage, we resize each image to $1024 \times 1024$ and adopt PhotoMetricDistortion augmentation, following \cite{2023HFT,2023DiffBEV}.
We use the AdamW \cite{kingma2014adam} optimizer with a total batch size of 64 for 40k iterations. The initial learning rate is set as 4e-4 for nuScenes \cite{2020nuscenesdataset} and 1e-4 for Argoverse \cite{2019argoversedataset}, which gradually increases in the first 1,500 iterations with the warm-up strategy and linearly decreases to 0 in the subsequent iterations.
We set hidden dimension $C_T = 512$. The transformer decoder in \secref{BCVT} consists of 2 layers and each layer has 4 attention heads.
The length of temporal fusion window $N_{his}$ is set as 2 in \secref{TempFusion} based on the ablation studies in \tabref{ablation_etf}. 
The resolution of BEV features in \secref{BCVT} and \secref{TempFusion} is $98 \times 100$ with each pixel converting $0.5m$. 
All experiments are conducted on 4 NVIDIA Tesla A100 GPUs.

\begin{table*}[t]
\centering{
\setlength\tabcolsep{6pt}

\caption{Comparisons with state-of-the-art methods on the ArgoVerse dataset. }
\label{argoverse benchmark}
\renewcommand{\arraystretch}{1.1}
\begin{tabular}{c|cccccccc||c}
\hline 
Methods   & Drivable      & Vehicle       & Pedestrian         & Large vehicle & Bicycle      & Bus           & Trailer       & Motor        & mIoU          \\ \hline \hline
PON \cite{2020PON}       & 65.7          & 27.7          & 6.6           & 8.1                & 0.3          & 19.9          & 16.5          & 0.2          & 18.1          \\
PYVA \cite{2021PYVA}      & 82.6          & 36.1          & 3.5           & 15.2               & 0.4          & 0.3           & \textbf{62.3} & 0.2          & 25.1          \\
BEVStitch \cite{2020bevstitch}    & 79.8          & 28.2          & 4.8           & 11.4               & 0.0          & 22.5          & 1.1           & 0.4          & 18.5          \\
TIM  \cite{2022TIM}     & 75.9          & 35.8          & 5.7           & 14.9               & 3.7          & 30.2          & 12.2          & 2.6          & 22.6          \\
HFT  \cite{2023HFT}     & 79.8          & 41.3          & \textbf{9.2}          & 26.5               & 8.4 & \textbf{36.5} & 39.3          & \textbf{7.6} & 31.0          \\ \hline 
Ours  & \textbf{86.5} & \textbf{43.7} & 8.8 & \textbf{32.0}      & \textbf{29.4}          & 26.1          & 55.0          & 0.0          & \textbf{35.2}\\ \hline
\end{tabular}
}
\end{table*}

\begin{figure*}
	\begin{center}
		\includegraphics[width=0.9\linewidth]{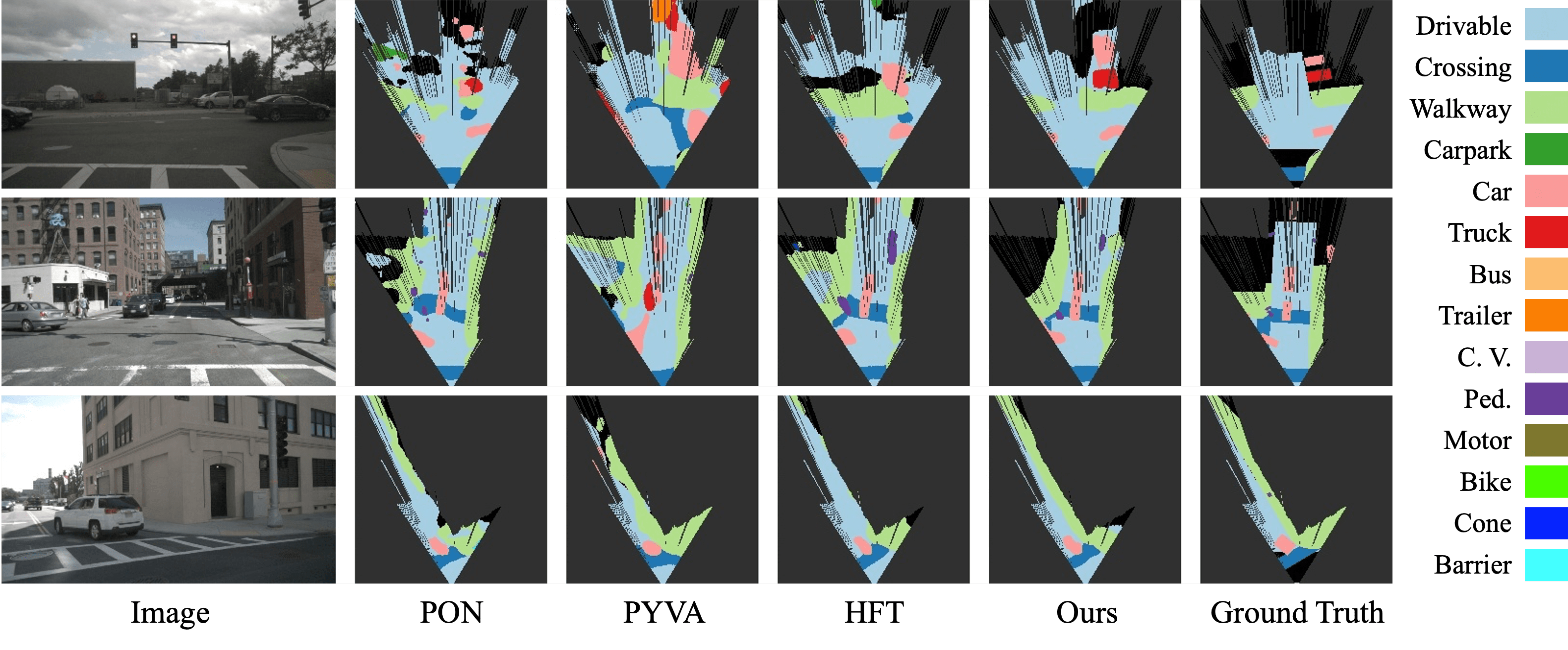}
		\caption{Qualiative results on the nuScenes dataset.
		}\label{fig:vis}
	\end{center}
\end{figure*}

\subsection{Comparison with State-of-the-art}
\textbf{Comparisons on nuScenes.} 
As shown in \tabref{nuscene benchmark}, we compare our approach on nuScenes benchmark with 10 state-of-the-art methods. 
It is worthy to notice that our approach achieves impressive performance on both static layout and dynamic objects, and outperforms previous state-of-the-art methods by a large margin, \eg, 3.5\% improvement compared with TIM \cite{2022TIM}.
As nuScenes is a more challenging benchmark for BEV segmentation tasks due to the greater diversity, the performance aforementioned demonstrates the superiority of our proposed modules.

\textbf{Comparisons on Argoverse.} In \tabref{argoverse benchmark}, we compare our approach on Argoverse benchmark with 5 state-of-the-art methods. We report the IoU of each class and mIoU for detailed comparisons on the commonly used backbone ResNet-50.
Notably, our approach achieves the best performance 35.2\%, and outperforms previous state-of-the-art HFT \cite{2023HFT} by a clear margin of 4.2\%.

\subsection{Performance Analysis}
\textbf{Ablation Studies.} 
To evaluate the effectiveness of each key component in our proposed approach, we reconstruct our model with different ablation factors in \tabref{ablation}. We first construct our baseline in the first row, employing ResNet-50 as the backbone and two-layer MLP as the view transformation module, similar to \cite{2020VPN,2022hdmapnet}, with weighted binary cross entropy loss and uncertain loss \cite{2020PON}. As shown in \tabref{ablation}, our proposed modules make a steady improvement on the high baseline performance, which demonstrates the effectiveness and necessity of the proposed modules.

\textbf{Effects of self-calibrated cycle view transformation.}
As shown in \tabref{ablation}, only adopting our proposed self-calibrated cycle view transformation (SCVT) can notably improve the performance,\ie, 9.7\% on total categories, which demonstrates the effectiveness of the cycle view transformation scheme. Besides, our proposed column-wise transformer decoder structure effectively boosts the performance from 25.7\% to 29.2\% compared with the MLP structure (\ie, the implementation of 0 decoder layer) as shown in \tabref{ablation_scvt}. Qualitative results of our approach in \figref{fig:vis} also reveal an advantage in clarity and completeness due to the focus on the BEV-relevant areas in view transformation.

\textbf{Effects of ego-motion-based temporal fusion.}
As shown in \tabref{ablation}, our proposed ego-motion-based temporal fusion module (ETF) could effectively improve the performance on both baseline and SCVT, especially for static layout categories. Furthermore, the static layout performance tends to increase with more history BEV features aggregated in the temporal fusion process as shown in \tabref{ablation_etf}, which demonstrates that our temporal fusion module benefits more from static spatiotemporal structure consistency. Qualitative results of our approach in \figref{fig:vis} also show an advantage in spatial continuity and layout integrity.


\section{CONCLUSION}
In this paper, we propose a novel FocusBEV framework for birds-eye-view segmentation tasks. 
Firstly, we propose a self-calibrated cycled view transformation module to suppress the BEV-agnostic image areas and focus on BEV-relevant areas in view transformation. Then an ego-motion-based temporal fusion module is adopted to exploit the spatiotemporal structure consistency in BEV space with a memory bank. Finally, we introduce an occupancy-agnostic IoU loss to address both semantic and positional uncertainties on the BEV plane.
Experimental evidence reveals that our proposed approach achieves new state-of-the-art on two popular benchmarks, \ie, 29.2\% on nuScenes and 35.2\% on Argoverse.











\bibliographystyle{IEEEtran}
\bibliography{IEEEabrv,myrefs}


\end{document}